\documentclass[journal]{IEEEtran}

\usepackage{amssymb,amsmath,amsthm}
\usepackage{graphicx}
\usepackage[ruled,noend]{algorithm2e}
\usepackage{url}
\usepackage{color}

\ifCLASSINFOpdf
\else
\fi
\begin{document}
%
% paper title
% Titles are generally capitalized except for words such as a, an, and, as,
% at, but, by, for, in, nor, of, on, or, the, to and up, which are usually
% not capitalized unless they are the first or last word of the title.
% Linebreaks \\ can be used within to get better formatting as desired.
% Do not put math or special symbols in the title.
\title{A Sparse Non-negative Matrix Factorization Framework for Identifying Functional Units of Tongue Behavior from MRI}

\author{Jonghye Woo,~\IEEEmembership{Member,~IEEE}, Jerry L. Prince,~\IEEEmembership{Fellow,~IEEE}, Maureen Stone, Fangxu Xing, Arnold D. Gomez, Jordan R. Green, Christopher J. Hartnick, Thomas J. Brady, Timothy G. Reese, Van J. Wedeen, Georges El Fakhri,~\IEEEmembership{Fellow,~IEEE}
       
% <-this % stops a space
\IEEEcompsocitemizethanks{\IEEEcompsocthanksitem Part of this work was presented at the MICCAI 2014 and Acoustical Society of America in 2017. Jonghye Woo, Fangxu Xing, Thomas J. Brady, Timothy R. Reese, Van J. Wedeen, and Georges El Fakhri are with Department of Radiology, Massachusetts General Hospital and Harvard Medical School. Maureen Stone is with the University of Maryland Dental School. Jordan R. Green is with MGH Institute of Health Professions. Christopher J. Hartnick is with Massachusetts Eye and Ear Infirmary and Harvard Medical School. Arnold Gomez and Jerry L. Prince are with Department of Electrical and Computer Engineering at Johns Hopkins University. 

©2018 IEEE. Personal use of this material is permitted. Permission from IEEE must be obtained for all other uses, including reprinting/republishing this material for advertising or promotional purposes, collecting new collected works for resale or redistribution to servers or lists, or reuse of any copyrighted component of this work in other works.\protect\\
% note need leading \protect in front of \\ to get a newline within \thanks as
% \\ is fragile and will error, could use \hfil\break instead.
E-mail: jwoo@mgh.harvard.edu}% <-this % stops an unwanted space
} %\thanks{Manuscript received xx, 2005}}

\markboth{IEEE Trans on Medical Imaging, 2018}%
{Shell \MakeLowercase{\textit{et al.}}: Bare Demo of IEEEtran.cls for Computer Society Journals}

\IEEEtitleabstractindextext{%
\begin{abstract}
Muscle coordination patterns of lingual behaviors are synergies generated by deforming local muscle groups in a variety of ways. Functional units are functional muscle groups of local structural elements within the tongue that compress, expand, and move in a cohesive and consistent manner. Identifying the functional units using tagged-Magnetic Resonance Imaging (MRI) sheds light on the mechanisms of normal and pathological muscle coordination patterns, yielding improvement in surgical planning, treatment, or rehabilitation procedures. Here, to mine this information, we propose a matrix factorization and probabilistic graphical model framework to produce building blocks and their associated weighting map using motion quantities extracted from tagged-MRI. Our tagged-MRI imaging and accurate voxel-level tracking provide previously unavailable internal tongue motion patterns, thus revealing the inner workings of the tongue during speech or other lingual behaviors. We then employ spectral clustering on the weighting map to identify the cohesive regions defined by the tongue motion that may involve multiple or undocumented regions. To evaluate our method, we perform a series of experiments. We first use two-dimensional images and synthetic data to demonstrate the accuracy of our method. We then use three-dimensional synthetic and \textit{in vivo} tongue motion data using protrusion and simple speech tasks to identify subject-specific and data-driven functional units of the tongue in localized regions.
\end{abstract}

% Note that keywords are not normally used for peer-review papers.
}
% make the title area
\maketitle

\IEEEdisplaynontitleabstractindextext

\IEEEpeerreviewmaketitle

%\IEEEraisesectionheading{\section{Introduction}\label{sec:introduction}}
\section{Introduction}

Finding a suitable representation of high-dimensional and complex data for various tasks, such as clustering and topic modeling, is a fundamental challenge in many areas such as computer vision, machine learning, data mining, and medical image analysis. Non-negative matrix factorization (NMF)~\cite{Lee_1999}, an unsupervised generative model, is a class of techniques to capture a low-dimensional representation of a dataset suitable for a clustering interpretation~\cite{deep_nmf}; it also has been used to transform different features to a common class. In this work, we are interested in modeling the tongue's underlying behaviors using NMF, since non-negative properties of NMF are akin to the physiology of the tongue as reflected in the matrix decomposition process. Since its introduction by Lee and Seung~\cite{Lee_1999}, special attention has been given to NMF and its variants involving sparsity because of its ability to uncover a parts-based and interpretable representation. Specifically, NMF with a sparsity constraint operates on input matrices whose entries are non-negative, thus allowing us to model a data matrix as sparse linear combinations of a set of basis vectors (or building blocks). NMF with a sparsity constraint only allows non-negative combinations of building blocks. This is analogous with the analysis of muscle synergies~\cite{Bizzi_2008,Kelso_2009} because a complex set of non-negative activations of tongue muscles generate the complexity and precision of the tongue's voluntary and involuntary movements.

The human tongue is one of the most structurally and functionally complex muscular structures in our body system, comprising orthogonally oriented and highly inter-digitated muscles~\cite{kier_1985}. The human tongue is innervated by over 13,000 hypoglossal motoneurons~\cite{Stone_2004,Kusky_1995}. A complex set of neural activations of tongue muscles allows to deform localized regions of the tongue in this complex muscular array. Muscle coordination patterns of the tongue are synergies produced by deforming local muscle groups---i.e., \textit{functional units}~\cite{Stone_2004,Kelso_2009}. Functional units can be thought of as an intermediate structure, which links muscle activity to surface tongue geometry, and which also exhibits cohesive and consistent motion in the course of specific tasks. Therefore, determining functional units and understanding the mechanisms by which muscles coordinate to generate target movements can provide new insights into motor control strategy in normal, pathological, and adapted tongue movements after surgery or treatment. This will, in turn, lead to improvement in surgical planning, treatment, or rehabilitation procedures. However, to date, the mechanisms of the muscle coordination patterns and the relationship between tongue structure and function have remained poorly understood because of the greater complexity and variability of both muscular structures and their interactions.

Understanding the subject/task-specific tongue's functional organization requires a map of the functional units of the tongue during the oromotor behaviors using medical imaging. In particular, recent advances in medical imaging and associated data analysis techniques have permitted the non-invasive imaging and visualization of tongue structure and function in a variety of ways. Magnetic resonance imaging (MRI) technologies have been widely used, demonstrating that a large number of tongue muscles undergo highly complex deformations during speech and other lingual behaviors. For example, the ability to image non-invasively using MRI has allowed to capture both surface motion of the tongue via cine- or real-time MRI~\cite{Stone_2001, Bresch_2008, Fu_2015} and internal tissue point motion of the tongue via tagged-MRI (tMRI)~\cite{Vijay_2007}. In addition, high-resolution MRI and diffusion MRI~\cite{gaige2007, shinagawa2008tongue} have allowed to image the muscular and fiber anatomy of the tongue, respectively. With advances in computational anatomy, a vocal tract atlas~\cite{Woo_2015,Stone_2016}, a representation of the tongue anatomy, has also been created and validated, which allows for investigating the relationship between tongue structure and function by providing a reference anatomical configuration to analyze similarities and variability of tongue motion.

This work is aimed at developing a computational approach for defining the subject/task-specific and data-driven functional units from tMRI and 4D (3D space with time) voxel-level tracking~\cite{Xing_2014} by extending our previous approach~\cite{Woo_2014}. In this work, we describe a refined algorithm including an advanced tracking method and graph-regularized sparse NMF (GS-NMF) to determine spatiotemporally varying functional units using protrusion and simple speech tasks and offer extensive validations on simulated tongue motion data. Since standard NMF formulation hinges on a Euclidean distance measure (i.e. Frobenius norm) or the Kullback-Leibler divergence, it fails to mine the intrinsic and manifold geometry of its data~\cite{Cai_2011}. Our method, therefore, makes use of both sparse and manifold regularizations to identify a set of optimized and geometrically meaningful motion patterns by promoting the computation of distances on a manifold from observed tongue motion data. Our formulation assumes that the tongue motion features live in a manifold space within a sparse NMF framework, thereby finding a low-dimensional yet interpretable subspace of the motion features from tMRI.

We present an approach to discovering the functional units using a matrix factorization and probabilistic graphical model framework with the following main contributions:
\begin{itemize}
  \item The most prominent contribution is to use voxel-level data-driven tMRI (1.875mm$\times$1.875mm$\times$1.875mm) methods incorporating internal tissue points to obtain subject-specific functional units of how tongue muscles coordinate to generate target observed motions. 
    \item Our tMRI imaging and 4D voxel-level tracking paradigm provide previously unavailable internal tongue motion patterns. The proposed approach is scalable to a variety of motion features derived from motion trajectories to characterize the coherent motion patterns. In this work, we consider the most representative features including displacements and angle from tMRI.
  \item This work applies a graph-regularized sparse NMF and probabilistic graphical model to the voxel-level motion data, allowing us to estimate the latent functional coherence, by learning simultaneously hidden building blocks and their associated weighting map from a set of motion features. 
  \item This work is aimed at identifying 3D cohesive functional units that involve multiple, and possibly undocumented and localized tongue regions unlike previous approaches that largely rely on the tracking of landmark points sparsely located on the fixed tongue surface such as the tip, blade, body, and dorsum.
\end{itemize}
The remainder of this paper is structured as follows. We review related work in Section II. Then, Section III shows the proposed method to identify functional units. The experimental results are provided and analyzed in Section IV. Section V presents a discussion, and finally Section VI concludes this paper.

\section{Related Work}\label{sec:previous}
In this section, we review recent work on NMF for clustering and functional units research that are closely related to our work.

\textbf{NMF for Clustering}. A multitude of NMF-based methods for unsupervised data clustering have been proposed over the last decade across various domains, ranging from video analysis to medical image analysis. In particular, the idea of using $L_{1}$ norm regularization (i.e., sparse NMF) for the purpose of clustering has been successfully employed~\cite{Kim_2008}. The sparsity condition imposed on the weighting matrix (or coefficient matrix) indicates the clustering membership. For example, Shahnaz et al.~\cite{Shahnaz_2006} proposed an algorithm for document clustering based on NMF, where the method was used to obtain a low-rank approximation of a sparse matrix while preserving natural data property. Wang et al.~\cite{Wang_2013} applied the NMF framework to gene-expression data to identify different cancer classes. Anderson et al.~\cite{Anderson_2014} presented an NMF-based clustering method to group differential changes in default mode subnetworks from multimodal data. In addition, Mo et al.~\cite{Mo_2012} proposed a motion segmentation method using an NMF-based method.  A key insight to use NMF for clustering purpose is that NMF is able to learn and discriminate localized traits of data with a better interpretation. Please refer to~\cite{Li_2013} for a detailed review on NMF for clustering purpose. 

\textbf{Speech Production}. Research on determining functional units during speech has a long history (see e.g., Gick and Stavness~\cite{Gick_2013} for a recent perspective). Determining functional units is considered as uncovering a ``missing link'' between speech movements primitives and cortical regions associated with speech production~\cite{Gick_2013}. In the context of lingual coarticulation, functional units can be seen as ``quasi-independent'' motions~\cite{Stone_2004}. A variety of different techniques have been used to tackle this problem. For instance, Stone et al.~\cite{Stone_1990} showed that four mid-sagittal regions including anterior, dorsal, middle, and posterior functioned quasi-independently using ultrasound and microbeam data. Green and Wang attempted to characterize functionally independent articulators from microbeam data using covariance-based analysis~\cite{Green_2003}. More recently, Stone et al.~\cite{Stone_2004} proposed an approach to determining functional segments using both ultrasound and tMRI. That work examined compression and expansion between the anterior and posterior part of the tongue and found that regions or muscles have functional segments influenced by phonemic constraints. Ramanarayanan et al.~\cite{Ramanarayan_2013} proposed to use a convolutive NMF method to identify motion primitives of the tongue using electromagnetic articulography (EMA). Being inspired by the aforementioned approaches, our work uses the augmented NMF framework in conjunction with the far richer 4D tMRI based voxel-level tracking to determine functional units of tongue behaviors.

\section{Methodology}\label{sec:method}
\subsection{Problem Statement}
Without loss of generality, let us first define the notations and definitions about tissue points and derived features used throughout this work. Let us consider a set of $P$ internal points of the tongue tracked through tMRI, in which tracked points are used to derive motion features. We now define motion features at each voxel location, having $n$ scalar quantities, such as magnitude and angle of each point track, precisely tracked through $L$ time frames. Each tissue point is characterized by these scalar quantities, which are then used to cluster them into coherent motion patterns. We denote the position of the $p$-th tissue point at the $l$-th time frame as ($x^{p}_{l}$, $y^{p}_{l}$, $z^{p}_{l}$). The tongue motion can be expressed as a $3L$$\times$$P$ spatiotemporal feature matrix $\mathbf{M} = [\mathbf{m}_1 \cdots \mathbf{m}_P] \in \mathbb{R}^{3L\times P}$, where the $p$-th column is expressed as

\begin{equation}
\mathbf{m}_p = [x_1^p \cdots x_L^p \;\; y_1^p \cdots y_L^p \;\; z_1^p \cdots z_L^p ]^T.
\end{equation}

The problem of identifying the functional units is viewed as a clustering problem, since the functional units are localized tongue regions that exhibit characteristic and homogeneous motions. However, different from generic motion clustering problems in computer vision or machine learning, the tongue's function and physiology also need to be reflected and captured in our formulation. Thus, we opt to identify a permutation of the columns to build $[\left. {\mathbf{M}_1} \cdots  {\mathbf{M}_c ]}\right.$, where the submatrix $\mathbf{M}_{i}$ comprises tracks belonging to the $i$-th submotion, corresponding to the $i$-th functional unit. The motion quantities for each underlying muscle of the tongue are not completely independent. That is, our framework is data-driven and motion-specific at the sub-muscle level, which could be mapped from a subset of motion quantities from one or more muscles. These latent morphologies---i.e., functional units---inherent in higher dimensional datasets offer a sparse representation of the generative process behind the tongue's physiology for sub-muscle, a single or multiple muscles. To achieve this goal, the sparse NMF method is utilized to capture the latent morphologies from motion trajectories. The proposed method is described in more detail below; a flowchart is depicted in Figure 1.

%%%%%%%%%%%%%%%%%%%%%%%%%%%%%%%%%%%%%%%%%%%%%%%%%%%%%%%%
\def\FigureHeight{58mm}
\begin{figure}[h]
 \center{
 \begin{tabular}{c@{ }c}
   \includegraphics[trim=0mm 0mm 0mm
0mm,clip=true,height=\FigureHeight]{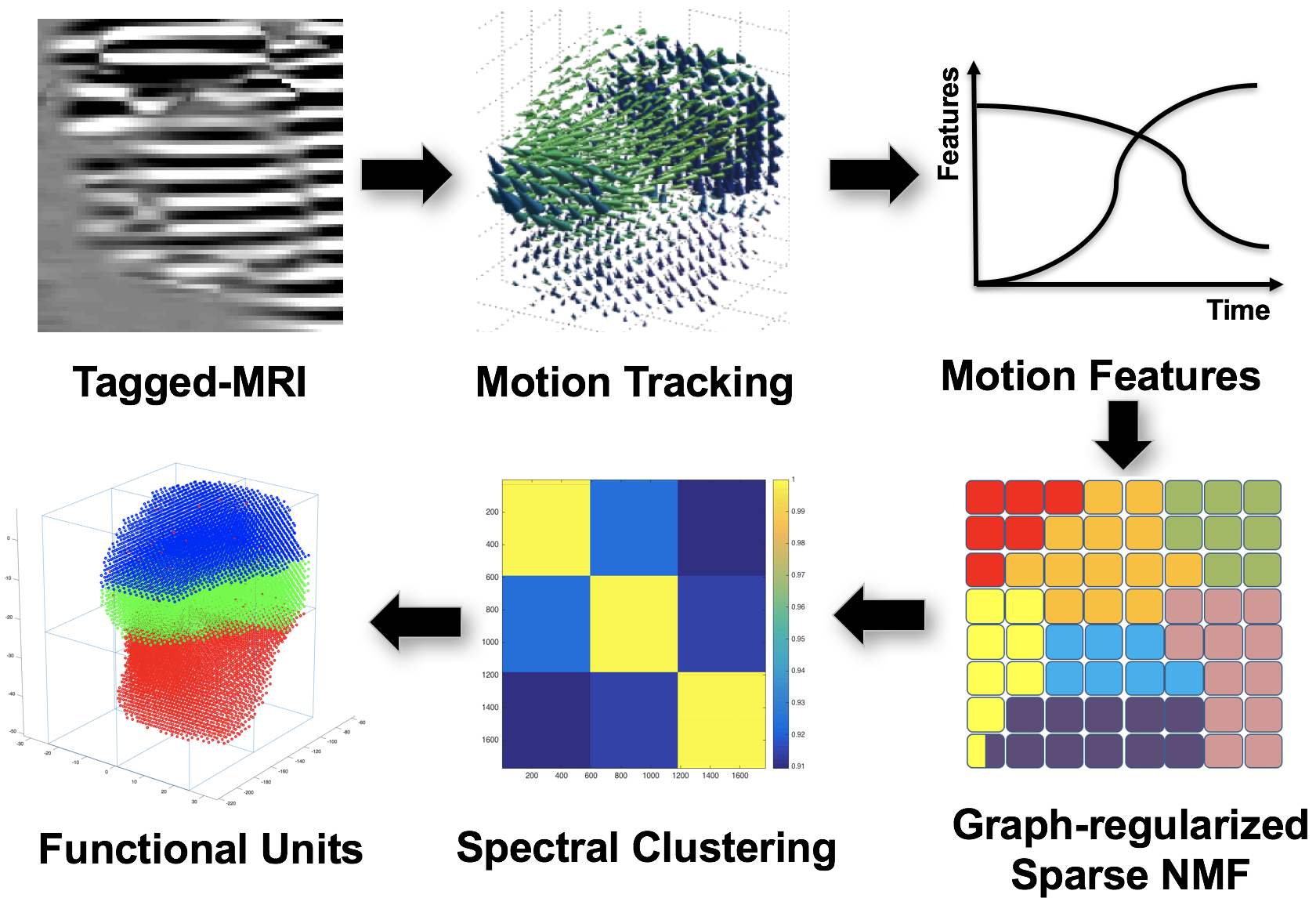}
& \hspace{0pt} 
\end{tabular}}
\caption{Overview of the proposed method}\label{fig:flow}
\end{figure}
%%%%%%%%%%%%%%%%%

\subsection{MR Data Acquisition and Motion Tracking}
\subsubsection{MR Data Acquisition} All MRI data are acquired on a Siemens 3.0 T Tim Treo system with 12-channel head and 4-channel neck coil. While a participant performs the protrusion task or simple speech tasks including ``a geese'' and ``a souk'' repeatedly in synchrony with metronome sound, the tMRI data are acquired using Magnitude Imaged CSPAMM Reconstructed images~\cite{NessAiver_2013}. All tMRI data have 26 time frames with a temporal resolution of 36 ms with no delay during a 1 second duration. The resolution is 1.875mm$\times$1.875mm$\times$6mm.

\subsubsection{Voxel-Level Motion Tracking from Tagged-MRI}
The phase vector incompressible registration algorithm (PVIRA)~\cite{pvira} is used to estimate deformation of the tongue from tMRI, yielding a sequence of voxel-level motion fields. The steps of the algorithm are as follows. First, although the input is a set of sparsely acquired tMRI slices, PVIRA uses cubic B-spline to interpolate these 2D slices into denser 3D voxel locations. Then a harmonic phase (HARP)~\cite{harp} filter is applied to produce HARP phase volumes from the interpolated result. Finally, PVIRA uses the iLogDemons method~\cite{ilogdemons} on these phase volumes. Specifically, we denote the phase volumes as $\Phi_{x}$, $\Theta_{x}$, $\Phi_{y}$, $\Theta_{y}$, $\Phi_{z}$, and $\Theta_{z}$, where $x$, $y$, and $z$ denote motion information from three cardinal directions usually contained in orthogonal axial, sagittal, and coronal tagged slices. The volume in the reference time frame is $\Phi$ and the volume in the deformed time frame is $\Theta$. The motion update vector field is derived from these phase volumes. At each voxel, the update vector is obtained by
\begin{equation}\label{eqnewdemonup}
\begin{aligned}
\delta \mathbf{v}(\mathbf{x}) &= \frac{\mathbf{v}_0(\mathbf{x})}{\alpha_1(\mathbf{x}) + \alpha_2(\mathbf{x})/S},
\end{aligned}
\end{equation}
where $S$ is the normalization factor determined by the mean squared value of the image voxel size~\cite{Nithiananthan_2009}.
 $\mathbf{v}_0(\mathbf{x})$, $\alpha_1(\mathbf{x})$, and $\alpha_2(\mathbf{x})$ are defined by
\begin{equation}\label{eqnewdemonup}
\begin{aligned}
\mathbf{v}_0(\mathbf{x}) &= W(\Phi_{x}(\mathbf{x}) - \Theta_{x}(\mathbf{x}))(\nabla^* \Phi_{x}(\mathbf{x}) + \nabla^* \Theta_{x}(\mathbf{x})) \\
&+ W(\Phi_{y}(\mathbf{x}) - \Theta_{y}(\mathbf{x}))(\nabla^* \Phi_{y}(\mathbf{x}) + \nabla^* \Theta_{y}(\mathbf{x})) \\
&+ W(\Phi_{z}(\mathbf{x}) - \Theta_{z}(\mathbf{x}))(\nabla^* \Phi_{z}(\mathbf{x}) + \nabla^* \Theta_{z}(\mathbf{x})) \;, \\
\alpha_1(\mathbf{x}) &= ||\nabla^* \Phi_{x}(\mathbf{x}) + \nabla^* \Theta_{x}(\mathbf{x})||^2 + ||\nabla^* \Phi_{y}(\mathbf{x}) + \nabla^* \Theta_{y}(\mathbf{x})||^2 \\
&+ ||\nabla^* \Phi_{z}(\mathbf{x}) + \nabla^* \Theta_{z}(\mathbf{x})||^2 \;, \\
\alpha_2(\mathbf{x}) &= W(\Phi_{x}(\mathbf{x}) - \Theta_{x}(\mathbf{x}))^2 + W(\Phi_{y}(\mathbf{x}) - \Theta_{y}(\mathbf{x}))^2 \\
&+ W(\Phi_{z}(\mathbf{x}) - \Theta_{z}(\mathbf{x}))^2. \;
\end{aligned}
\end{equation}
$\mathbf{v}(\mathbf{x})$ is a stationary motion field that is used as an intermediate step towards the final motion field $\varphi(\mathbf{x})$. Wrapping of phase $W(\theta)$ is defined by
\begin{equation}\label{eqphasegrad}
W(\theta) = \mathrm{mod} (\theta + \pi, 2\pi) - \pi
\end{equation}
and the ``starred'' gradient is defined by
\begin{equation}\label{eqphasegrad}
\nabla^*\Phi(\mathbf{x}) :=
\left\{
\begin{array}{l}
\nabla\Phi(\mathbf{x}), \qquad \qquad \mathrm{if}\; |\nabla\Phi(\mathbf{x})| \leq |\nabla W(\Phi(\mathbf{x})+\pi)| \;\\
\nabla W(\Phi(\mathbf{x})+\pi),\; \mathrm{otherwise.}\;
\end{array}
\right.
\end{equation}
After all iterations are complete, the forward and inverse motion fields can be found by
\begin{equation}\label{eqphasegrad_}
\varphi(\mathbf{x}) = \exp (\mathbf{v}(\mathbf{x}))\,\,\,\mathrm{and}\,\, \varphi^{-1}(\mathbf{x}) = \exp(-\mathbf{v}(\mathbf{x})),
\end{equation}
and they are both incompressible and diffeomorphic, making both Eulerian and Lagrangian computations available for the following continuum mechanics operations. Once the motion field $\varphi(\mathbf{x})$ is estimated, the new voxel locations are mapped by applying the obtained motion field to the time frame under consideration.

\subsection{Extraction of Motion Features}
We first extract the motion features from PVIRA to characterize the coherent motion patterns. We use the magnitude and angle of each track as our motion features similar to~\cite{Cheriyadat_2009} as described as
\begin{equation}
m^{p}_{l} = \sqrt{(x^{p}_{l+1}-x^{p}_{l})^2 + (y^{p}_{l+1}-y^{p}_{l})^2 + (z^{p}_{l+1}-z^{p}_{l})^2}
\end{equation}
\begin{equation}
oz_l^p  = \frac{{x_{l+1}^p  - x_l^p }}
{{\sqrt {(x_{l + 1}^p  - x_l^p )^2  + (y_{l + 1}^p  - y_l^p )^2 } }} + 1
\end{equation}
\begin{equation}
ox_l^p  = \frac{{y_{l+1}^p  - y_l^p }}
{{\sqrt {(y_{l+1}^p  - y_l^p )^2  + (z_{l+1}^p-z_l^p )^2 } }}+ 1
\end{equation}
\begin{equation}
oy_l^p  = \frac{{z_{l+1}^p - z_l^p }}
{{\sqrt {(z_{l+1}^p  - z_l^p )^2  + (x_{l+1}^p-x_l^p )^2 } }}+ 1,
\end{equation}
where $m^{p}_{l}$ is the magnitude of each track and $oz_l^p$, $ox_l^p$, and $oy_l^p$ denote the cosine of the angle after projecting two consecutive adjacent point tracks in the $z$, $x$, and $y$ axes plus one, respectively. We rescale all the features into the range of 0 to 10 for each feature to be comparable and to satisfy the non-negative constraint of the NMF formulation.

For further clustering, all the motion features are combined into a single $4(L-1)\times P$ non-negative matrix $\mathbf{U} =
[\mathbf{u}_1,...,\mathbf{u}_n] \in \mathbb{R}^{m\times n}_+$, where the $p$-th column is given by 

\begin{IEEEeqnarray}{rCl}
\begin{gathered}
\mathbf{u}_p = [m_1^p \cdots m_{L-1}^p \;\; oz_1^p \cdots oz_{L-1}^p \;\; ox_1^p \cdots ox_{L-1}^p oy_1^p \\ \cdots oy_{L-1}^p ]^T. 
\end{gathered}
\end{IEEEeqnarray}
Since each value of the features is non-negative, they can be input to our framework.

\subsection{Graph-regularized Sparse NMF}
\subsubsection{NMF} 
Using a non-negative feature matrix $\mathbf{U}$ built by the motion quantities described above and $K \leq \rm{min}$$(m, n)$, let $\mathbf{V} = [v_{ik}] \in \mathbb{R}^{m\times K}_+$ be the \textit{building blocks} and $\mathbf{W} = [w_{kj}] \in \mathbb{R}^{K\times n}_+$ be the \textit{weighting map}, respectively. The objective of NMF is to produce two output non-negative matrices (i.e., $\mathbf{U} \approx \mathbf{VW}$): (1) spatial building blocks, which can change over time and (2) a weighting map of those blocks, which weights them over time. We define NMF using the Frobenius norm to minimize the reconstruction error between $\mathbf{U}$ and $\mathbf{VW}$~\cite{Lee_1999} as given by
\begin{equation}
\mathcal{E}(\mathbf{V, W}) =  \left\| {\mathbf{U} - \mathbf{VW}} \right\|_F^2 = \sum\limits_{i,j} {\left( {u_{ij}  - \sum\limits_{k = 1}^K {v_{ik} w_{kj} } } \right)}^2
\end{equation}
where $\left\|  \cdot  \right\|_F$ is the matrix Frobenius norm. A popular choice to solve this is to use the multiplicative update rule~\cite{Lee_1999}:
\begin{equation}
\mathbf{V} \leftarrow \mathbf{V} \circ \frac{\mathbf{{UW}}^T}{\mathbf{VWW}^T}  
\end{equation}
\begin{equation}
\mathbf{W} \leftarrow \mathbf{W} \circ \frac{\mathbf{V}^T\mathbf{U}} {\mathbf{V}^T\mathbf{VW}}, \hfill \\ 
\end{equation}
where the $\circ$ operator denotes element-wise multiplication and the division is element-wise as well.

\subsubsection{Sparsity Constraint} Once we obtain the building blocks and their weighting map, the identified weighting map is used to identify the cohesive region, which in turn exhibits functional units. Since there could be many possible weighting maps to generate the same movements, a sparsity constraint is incorporated into the weighting map so that the obtained functional units represent optimized and simplest tongue behaviors to generate target movements. This is also analogous with the current wisdom on phonological theories~\cite{Browman_1995}. In essence, we use the sparsity constraint to encode the high-dimensional and complex tongue motion data using a small set of active (or non-negative) components so that the obtained weighting map is simple and easy to interpret. Within the sparse NMF framework, a fractional regularizer using the $L_{1/2}$ norm has been shown to provide superior performance to the $L_1$ norm regularization by producing sparser solutions~\cite{Qian_2009}. Thus, in this work, we use the $L_{1/2}$ sparsity constraint in our NMF framework, which is given by 
\begin{equation}
\mathcal{E}(\mathbf{V, W}) =  {\frac{1}
{2} \left\| {\mathbf{U} - \mathbf{VW}} \right\|_F^2 + \eta \left\| \mathbf{W} \right\|_{1/2}}.
\end{equation}
Here, the parameter $\eta \geqslant 0$ is a weight associated with the sparseness of $\mathbf{W}$ and $\left\| \mathbf{W} \right\|_{1/2}$ is expressed as 
\begin{equation}
\left\| \mathbf{W} \right\|_{1/2} = \left( {\sum\limits_{i = 1}^k
  {\sum\limits_{j = 1}^n {w_{ij}^{1/2} } } } \right)^{2} \,. 
\end{equation}

\subsubsection{Manifold Regularization} Despite the high-dimensional configuration space of human motions, many motions lie on low-dimensional and non-Euclidean manifolds~\cite{Elgammal_2008}. NMF with the sparsity constraint described above, however, produces a weighting map using a Euclidean structure in the high-dimensional space. Therefore, the intrinsic and geometric relation between motion quantities cannot be captured faithfully in the sparse NMF formulation. Thus, we added a manifold regularization to capture the intrinsic geometric structure similar to~\cite{Cai_2011, Yang_2011, Lu_2013}. The manifold regularization additionally preserves the local geometric structure. Our final cost function using both regularizations is then given by
\begin{equation}
\mathcal{E}(\mathbf{V, W}) = \frac{1}{2}{\left\| {\mathbf{U} - \mathbf{VW}} \right\|_F^2 + \frac{\lambda }{2}\mathrm{Tr}(\mathbf{WLW}^T) + \eta \left\| \mathbf{W} \right\|_{1/2} } % \right],
\label{eq:eq10}
\end{equation}
where $\lambda$ represents a weight associated with the manifold regularization, Tr($\cdot$) represents the trace of a matrix, $\mathbf{Q}$ denotes a heat kernel weighting, $\mathbf{D}$ denotes a diagonal matrix, where $\mathbf{D}_{jj}  = \sum\limits_l {\mathbf{Q}_{jl}}$ and $\mathbf{L=D-Q}$, which is the graph Laplacian.

\subsubsection{Minimization} Due to the non-convex cost function in Eq.~(\ref{eq:eq10}), a multiplicative iterative method is adopted akin to that used in~\cite{Lu_2013}. Let $\mathbf{\Psi} = \left[\psi_{mk} \right]$ and $\mathbf{\Phi} = \left[\phi_{kn} \right]$ be Lagrange multipliers subject to $v_{mk} \geq 0$ and $w_{kn} \geq 0$, respectively. We solve this using the Lagrangian by the definition of the Frobenius norm, $\left\| \mathbf{U} \right\|_F  = (\mathrm{Tr}(\mathbf{U}^T \mathbf{U}))^{1/2}$, and matrix algebra given by 
\begin{equation}
\begin{gathered}
  \mathcal{L} = \frac{1}
{2}\mathrm{Tr}(\mathbf{UU}^T ) - \mathrm{Tr}(\mathbf{UW}^T \mathbf{V}^T ) + \frac{1}
{2}\mathrm{Tr}(\mathbf{VWW}^T \mathbf{V}^T ) \hfill \\
   + \frac{\lambda }
{2}\mathrm{Tr}(\mathbf{WLW}^T ) + \mathrm{Tr}(\mathbf{\Psi} \mathbf{V}^T ) + \mathrm{Tr}(\mathbf{\Phi} \mathbf{W}^T ) + \eta \left\| \mathbf{W} \right\|_{1/2}.  \hfill \\ 
\end{gathered} 
\end{equation}
The partial derivatives of $\mathcal{L}$ with respect to $\mathbf{V}$ and $\mathbf{W}$ are given by
\begin{equation}
  \frac{{\partial \mathcal{L}}}
{{\partial \mathbf{V}}} =  - \mathbf{UW}^T  + \mathbf{VWW}^T  + \mathbf{\Psi} 
\end{equation}
\begin{equation}
  \frac{{\partial \mathcal{L}}}
{{\partial \mathbf{W}}} =  - \mathbf{V}^T \mathbf{U} + \mathbf{V}^T \mathbf{VW} + \lambda \mathbf{WL} + \frac{\eta }
{2}\mathbf{W}^{-1/2} + \mathbf{\Phi}. 
\end{equation}
Finally, the update rule is found by using Karush-Kuhn-Tucker conditions---i.e.,
$\psi_{mk} v_{mk} = 0$ and $\phi_{kn} w_{kn} = 0$: 
\begin{equation}
  \mathbf{V} \leftarrow \mathbf{V} \circ \frac{\mathbf{{UW}}^T}{\mathbf{VWW}^T}
  \end{equation}
\begin{equation}
  \mathbf{W} \leftarrow \mathbf{W} \circ \frac{\mathbf{V}^T \mathbf{U} + \lambda \mathbf{WQ}}{\mathbf{V}^T \mathbf{VW} + \frac{\eta}
{2}\mathbf{W}^{ - 1/2} + \lambda \mathbf{WD}}. 
\label{eq:eq13}
\end{equation}
\subsection{Spectral Clustering}
Now that we obtain the building blocks and their associated weighting map in Eq.~(\ref{eq:eq13}), the widely used spectral clustering is applied to the weighting map which represents a measure of tissue point similarity. It has been reported that spectral clustering outperforms conventional clustering methods such as the K-means algorithm~\cite{Luxburg_2013}. We use spectral clustering to obtain the final clustering results, which in turn reveals the cohesive motion patterns. Of note, alternative clustering algorithms could be applied in this step.

More specifically, from the weighting map $\mathbf{W}$ as in Eq.~(\ref{eq:eq13}), an affinity matrix $\mathbf{A}$ is constructed: 
\begin{equation}
\mathbf{A}(i,j) = \exp \left( { - \frac{{\left\| {w(i) - w(j)} \right\|_2 }}
{\sigma }} \right),
\end{equation}
where $w(i)$ represents the $i$-th column vector of $\mathbf{W}$ and $\sigma$ is the scale. Within the graph cut framework, nodes in the graph are formed by the column vectors of $\mathbf{W}$, and the edge weights indicate the similarity $\mathbf{A}$ calculated between column vectors of $\mathbf{W}$. On the affinity matrix, we apply spectral clustering using a normalized cut algorithm~\cite{Shi_2000}. From a graph cut point of view, our method can be seen as identifying sub-graphs representing characteristic motions that are different from one another.

\subsection{Model Selection}

To achieve the best clustering quality, we need to select the optimal number $k$ of clusters, which however is a challenging task. Previous research on coarticulation examined a small number of tongue regions to assess the degrees of freedom of the vocal tract ranging from two (i.e., tip vs. body)~\cite{ohman_1967} to five units (i.e., front-to-back segments for genioglossus, verticalis, and transverse muscles)~\cite{Stone_2004}. In addition to this empirical knowledge about tongue anatomy and physiology, in this work, we further use a consensus clustering approach introduced by Monti et al.~\cite{Monti_2003} to estimate the optimal number of clusters within the NMF framework by examining the stability of the identified clusters. In brief, NMF may yield different decomposition results depending on different initializations~\cite{Brunet_2004, clusters_2015}. Therefore, the consensus clustering approach is based on the assumption that sample assignments to clusters would not change depending on the different runs. A connectivity matrix $C$ is constructed with entry $c_{ij}$=1 if samples $i$ and $j$ are clustered together, and $c_{ij}$=0 if samples are clustered differently. The consensus matrix ${\tilde C}$ is then constructed by taking the average over the connectivity matrices generated by different initializations. In this work, we select the number of runs (30 times) based on the stability of the consensus matrix. Finally, the dispersion coefficient $\rho$ of the consensus matrix ${\tilde C}$ is defined as
\begin{equation}
\rho  = \frac{1}{{{n^2}}}\sum\limits_{i = 1}^n {\sum\limits_{j = 1}^n {4{{({{\tilde c}_{ij}} - 0.5)}^2}}},
\end{equation}
where ${\tilde c}_{ij}$ and $n$ denote each entry of the matrix and the row and column size of the matrix, respectively. Since the entries of the consensus matrix represent the probability that samples $i$ and $j$ are clustered together, the dispersion coefficient, ranging between 0 and 1, represents the reproducibility of the assignments using different initializations. The optimal number of clusters is then chosen as the one that yields the maximal value.
%% Table 1 %%%%%%%%%%%%%%%%%%%%%%%%%%%%%%%%%%%%%%%%%%%%%%%%%%%%%%%
\begin{table}[h]
\caption{Clustering Performance}
\centering
%\scriptsize{ 
\begin{tabular}{c||c|c|c|c} \hline
 \hline 
  AC ($\%$) & K-means  & N-Cut  & GS-NMF-K & Our method \\
   \hline \hline
 COIL20 ($\mathcal{K}$=20) & 60.48$\%$ & 66.52$\%$  & 83.75$\%$ & \textbf{85.00}$\%$ \\ 
 PIE ($\mathcal{K}$=68) &  23.91$\%$ & 65.91$\%$ & 79.90$\%$ & \textbf{84.13}$\%$  \\ 
Tongue ($\mathcal{K}$=7) &  98.41$\%$ & 99.71$\%$ & 98.51$\%$ & \textbf{99.92}$\%$ \\
 Tongue ($\mathcal{K}$=8) & 98.72$\%$ & 99.85$\%$ & 99.14$\%$ & \textbf{99.89}$\%$ \\
 \hline 
\end{tabular}\label{table:2d}
\end{table}
%%%%%%%%%%%%%%%%%
\section{Experimental Results}
In this section, we describe the qualitative and quantitative evaluation to validate the proposed method. We used 2D image and tongue data, and 3D simulated tongue motion data to demonstrate the accuracy of our approach. Furthermore, 3D \textit{in vivo} human tongue motion datasets were used to identify functional units during protrusion and simple speech tasks. All experiments were performed on an Intel Core i5 with a clock speed of 3.1 GHz and 16GB memory. The mean computation times for the clustering algorithm for ``a souk,'' ``a geese,'' and the protrusion task were 18.3, 21.2, and 10.4 secs, respectively.

\subsection{Experiments Using 2D Image and Synthetic Motion Data}
 
 We used three 2D datasets to validate our method. The first two datasets include the COIL20 image database and the CMU PIE database, each having 20 and 68 labels, respectively. As our third dataset, we used 2D synthetic tongue motion data containing a displacement field from two time frames as shown in Figure~\ref{fig:synthetic_tongue}, where we tested 7 and 8 labels. Our method was compared against K-means clustering, a normalized cut method (N-Cut)~\cite{Shi_2000}, and graph-regularized sparse NMF with K-means clustering (GS-NMF-K)~\cite{Cai_2011}, given the known ground truth and the number of labels. To measure the accuracy,  the clustering accuracy (AC) was used similar to~\cite{Cai_2011}. Table~\ref{table:2d} lists the accuracy measure, showing that our method performs better than other methods. In addition, the $L_{1/2}$ and $L_{1}$ norms were compared experimentally, demonstrating that the $L_{1/2}$ norm performed better. In our experiments, we chose $\eta$=70, $\lambda$=100, and $\sigma$ = 0.01 for the COIL dataset,  $\eta$=160, $\lambda$=100, and $\sigma$ = 0.01 for the PIE dataset, and $\eta$=100, $\lambda$=800, and $\sigma$ = 0.06 for tongue datasets. We chose the parameters empirically that yielded the maximal performance.

%%%%%%%%%%%%%%%%%%%%%%%%%%%%%%%%%%%%%%%%%%%%%%%%%%%%%%%%
\def\FigureHeight{66mm}
\begin{figure}[h]
 \center{
 \begin{tabular}{c@{ }c}
   \includegraphics[trim=0mm 0mm 0mm
0mm,clip=true,height=\FigureHeight]{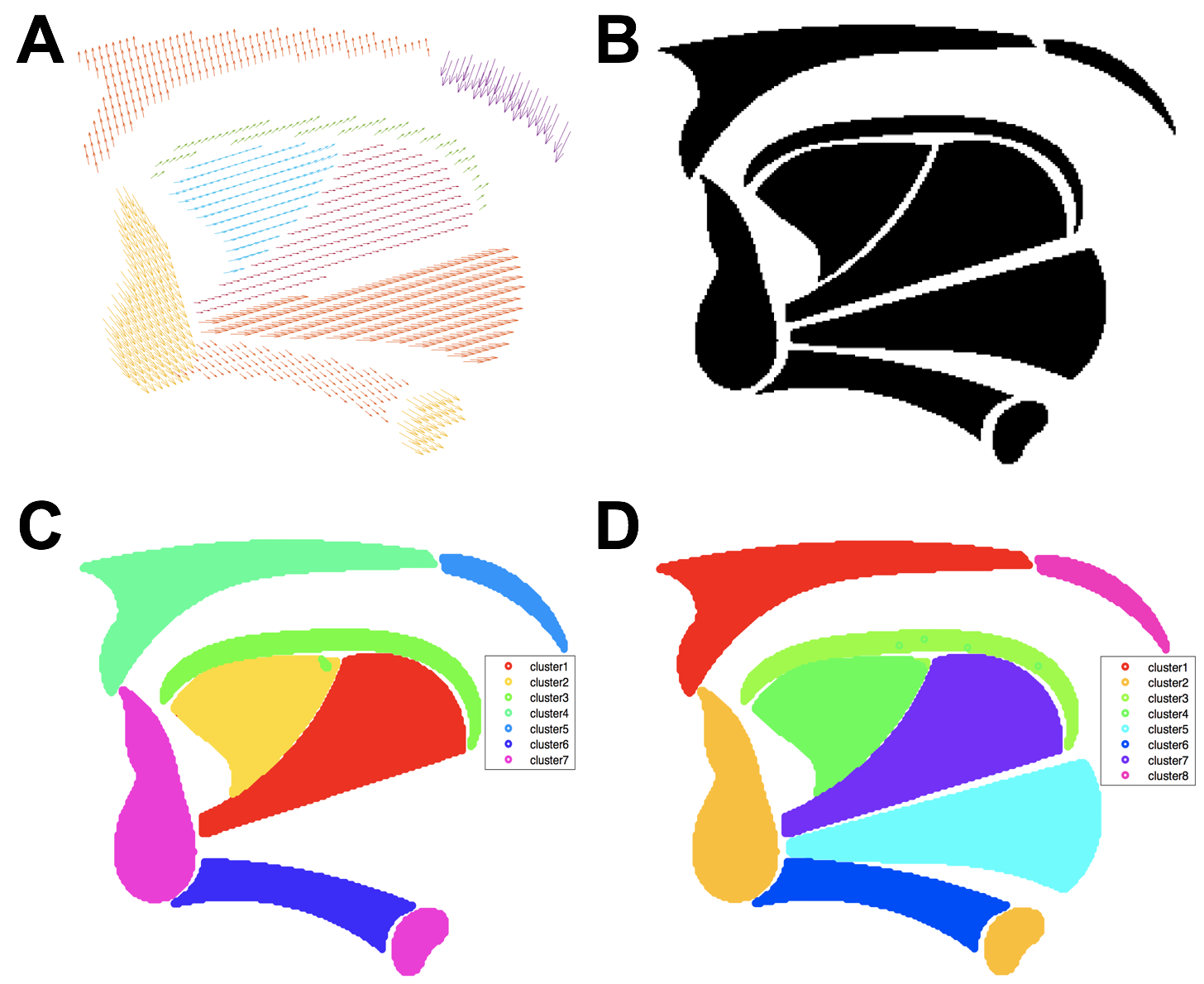}
& \hspace{0pt} 
\end{tabular}}
\caption{Illustration of 2D synthetic tongue motion simulation results: (A) a quiver plot of the synthetic displacement field for each region ($\mathcal{K}$=8), (B) ground truth labels ($\mathcal{K}$=8), (C) clustering results of 7 regions using our method, and (D) clustering results of 8 regions using our method.}\label{fig:synthetic_tongue}
\end{figure}
%%%%%%%%%%%%%%%%%

%%%%%%%%%%%%%%%%%%%%%%%%%%%%%%%%%%%%%%%%%%%%%%%%%%%%%%%%
\def\FigureHeight{82mm}
\begin{figure}[h]
 \center{
 \begin{tabular}{c@{ }c}
   \includegraphics[trim=0mm 0mm 0mm
0mm,clip=true,height=\FigureHeight]{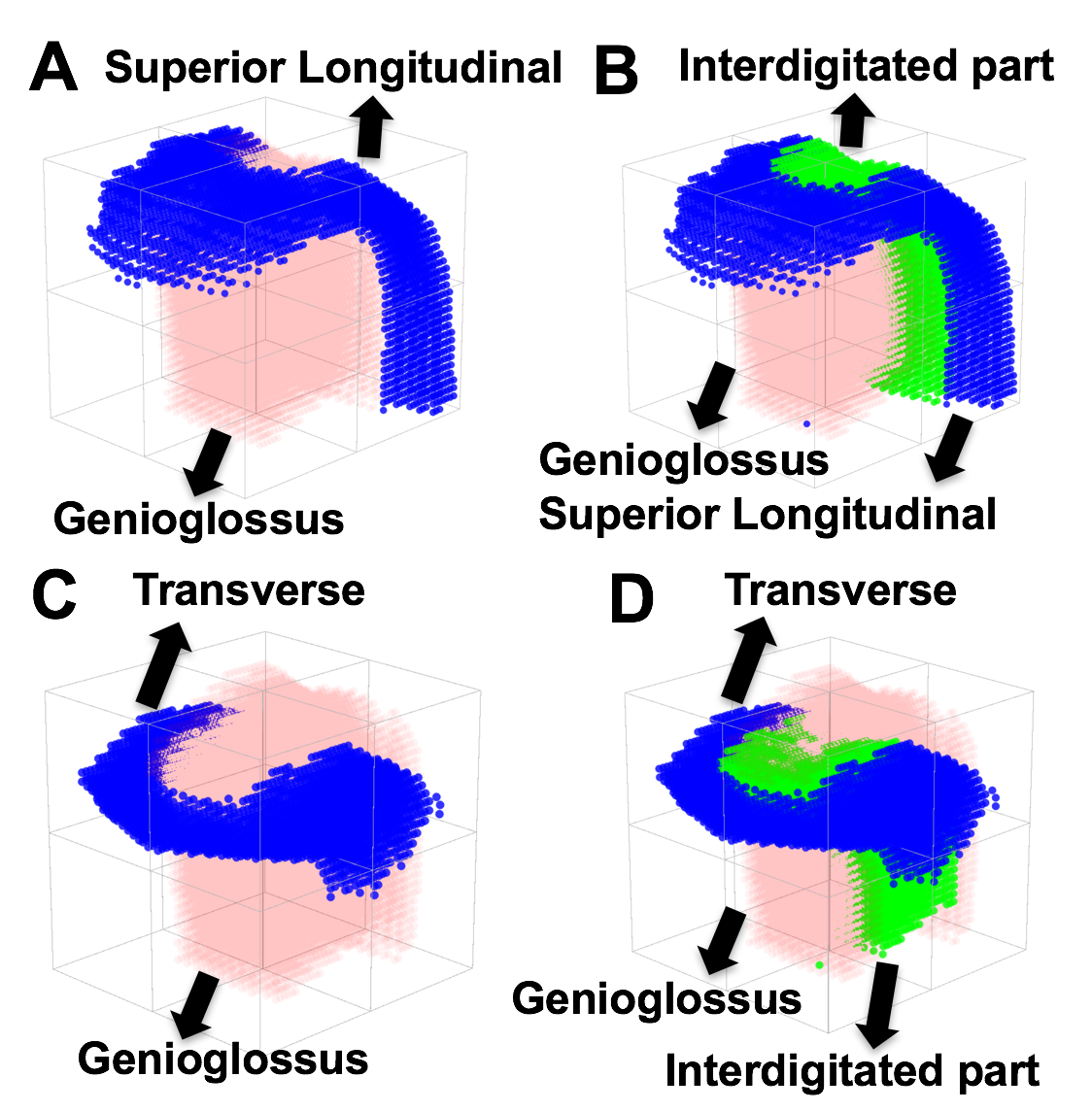}
& \hspace{0pt} 
\end{tabular}}
\caption{Illustration of clustering results for 3D synthetic tongue motion simulations: (A) translation plus rotation without interdigitated regions (2 clusters), (B) rotations with interdigitated regions (3 clusters), (C) translation plus rotation without interdigitated regions (2 clusters), and (D) rotations with interdigitated regions (3 clusters). It is noted that our approach identified each label accurately as visually assessed.}\label{fig:synthetic}
\end{figure}
%%%%%%%%%%%%%%%%%

%% Table 1 %%%%%%%%%%%%%%%%%%%%%%%%%%%%%%%%%%%%%%%%%%%%%%%%%%%%%%%
\begin{table}[h]
\caption{Clustering Performance}
\centering
%\scriptsize{ 
\begin{tabular}{c||c|c|c|c} \hline \hline
 AC ($\%$) & K-means  & N-Cut  & GS-NMF-K & Our method \\
 \hline \hline
 A ($\mathcal{K}$=2) & 91.72$\%$ & 75.84$\%$ & 75.64$\%$ & \textbf{94.79}$\%$ \\ 
 B ($\mathcal{K}$=3) & 94.83$\%$ & 95.51$\%$ & 96.16$\%$ & \textbf{96.44}$\%$  \\ 
 C ($\mathcal{K}$=2) & 87.91$\%$ & \textbf{100}$\%$ & 86.43$\%$ & \textbf{100}$\%$ \\
 D ($\mathcal{K}$=3) & 81.24$\%$ & \textbf{99.99}$\%$ & 85.31$\%$ & \textbf{99.99}$\%$ \\
 \hline 
\end{tabular}\label{table:3d}
\end{table}
%%%%%%%%%%%%%%%%%

\subsection{Experiments Using 3D Synthetic Tongue Motion Data}
We also assessed the performance of the proposed method using four synthetic tongue motion datasets because no ground truth is available in our tongue motion data. We used a composite Lagrangian displacement field of individual muscle groups based on a tongue atlas~\cite{Woo_2015}. Each muscle group was defined by a mask volume with a value of 1 inside the muscle group, and zero elsewhere. Since the masks were known, it also provided ground truth labels to examine the accuracy of the output of our method. The first and second datasets used genioglossus (GG) and superior longitudinal (SL) muscles. Note that those two muscles interdigitate with each other. In the first dataset, the GG muscle was translated while the SL muscle was rotated $-0.1$ radians about the $x$ direction in the course of 11 time frames. The interdigitated part was included in the GG muscle. The clustering outcome (i.e., 2 clusters) using our method was displayed in Fig.~\ref{fig:synthetic}(A). In the second dataset, the same motion was used in the first dataset for the clustering, where the clustering outcome (i.e., 3 clusters) using our method was displayed in Fig.~\ref{fig:synthetic}(B). The third and fourth datasets used GG and transverse muscles. Note that those two muscles also interdigitate with each other. In the third dataset, the GG muscle was rotated $-0.1$ radians about the $x$ direction while the transverse muscle was translated in the course of 11 time frames. The clustering outcome (i.e., 2 clusters) using our method was displayed in Fig.~\ref{fig:synthetic}(C). The interdigitated part was included in the GG muscle. In the fourth dataset, the GG muscle was translated while the transverse muscle rotated $-0.1$ radians about the $x$ direction in the course of 11 time frames. The clustering outcome (i.e., 3 clusters) using our method was displayed in Fig.~\ref{fig:synthetic}(D). Table~\ref{table:3d} lists the AC measure, demonstrating that our method outperforms other methods. In our experiments, we chose $\eta$=100, $\lambda$=100, and $\sigma$ = 0.01 for A and C datasets, and $\eta$=100, $\lambda$=100, and $\sigma$ = 0.05 for B and D datasets.

%with and without interdigitated regions. The GG muscle was rotated $-0.1$ radians about the $x$ direction while the SL muscle was translated in Fig.~\ref{fig:synthetic}(A) in the course of 11 time frames. The GG muscle was rotated $-0.1$ radians about the $x$ direction while the SL muscle was rotated $0.1$ radians about the $x$ direction in Fig.~\ref{fig:synthetic}(B) in the course of 11 time frames. The third and fourth datasets were generated by applying the same composite Lagrangian displacement field to the GG and transverse muscles with and without interdigitated regions as in Fig.~\ref{fig:synthetic}(C) and (D), respectively. Fig.~\ref{fig:synthetic} showed the final clustering results using our method. We attempted to cluster each dataset into two (first and third datasets) and three (second and fourth datasets) distinct motions, respectively, where we obtained 100$\%$ clustering accuracy for all datasets when evaluated against the ground truth labels. 

\subsection{Experiments Using 3D \textit{In Vivo} Tongue Motion Data}
%% Table 1 %%%%%%%%%%%%%%%%%%%%%%%%%%%%%%%%%%%%%%%%%%%%%%%%%%%%%%%
\begin{table}[h]
\caption{Characteristics of \textit{in vivo} tongue motion data}
\centering
%\scriptsize{ 
\begin{tabular}{c||c|c|c} \hline
 Task & Protrusion & /s/-/u/ & /i/-/s/  \\
 \hline \hline
 Time frames & 1-13 & 10-17 & 15-26\\ 
 Number of clusters & 2-5 & 2-5 & 2-5 \\
 \hline 
\end{tabular}\label{table:subj}
\end{table}
%%%%%%%%%%%%%%%%%
3D \textit{in vivo} tongue motion data including protrusion and simple speech tasks such as ``a souk" and ``a geese" were used to identify functional units. One subject performed the protrusion task synchronized with metronome sound, and ten subjects performed ``a souk" and ``a geese" tasks using the same protocol. We used the same motion quantities described above as our input to the sparse NMF framework. Table~\ref{table:subj} lists the characteristics of our \textit{in vivo} tongue motion data including time frames analyzed and the number of clusters extracted based on the dispersion coefficient. We used $\eta$=100, $\lambda$=100, and $\sigma$ = 0.01 for our 3D \textit{in vivo} tongue datasets. Of note, we focused on 2-4 detected functional units in our interpretation and analysis below.

%%%%%%%%%%%%%%%%%%%%%%%%%%%%%%%%%%%%%%%%%%%%%%%%%%%%%%%%
\def\FigureHeight{39mm}
\begin{figure*}[h]
 \center{
 \begin{tabular}{c@{ }c}
   \includegraphics[trim=0mm 0mm 0mm
0mm,clip=true,height=\FigureHeight]{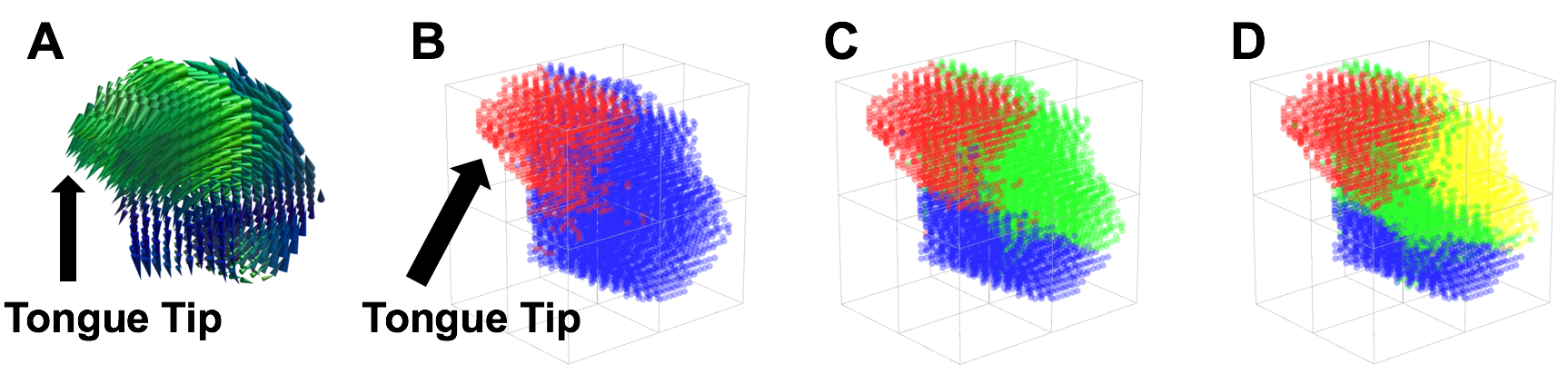}
& \hspace{0pt} 
\end{tabular}}
\caption{Illustration of identified functional units using our method during the tongue protrusion task, depicting (A) 3D Lagrangian displacement field, (B) detected functional units (2 clusters), (C) detected functional units (3 clusters), and (D) detected functional units (4 clusters). Please note that the displacement fields and clustering results are plotted relative to the first time frame representing the neutral tongue position. The cone size in (A) represents the magnitude of the displacement field, and red, green, and blue represent left--right, front--back, and up--down directions of the displacement field, respectively.}\label{fig:prot_functional_unit}
\end{figure*}
%%%%%%%%%%%%%%%%%

%%%%%%%%%%%%%%%%%%%%%%%%%%%%%%%%%%%%%%%%%%%%%%%%%%%%%%%%
\def\FigureHeight{79mm}
\begin{figure}[h]
 \center{
 \begin{tabular}{c@{ }c}
   \includegraphics[trim=0mm 0mm 0mm
0mm,clip=true,height=\FigureHeight]{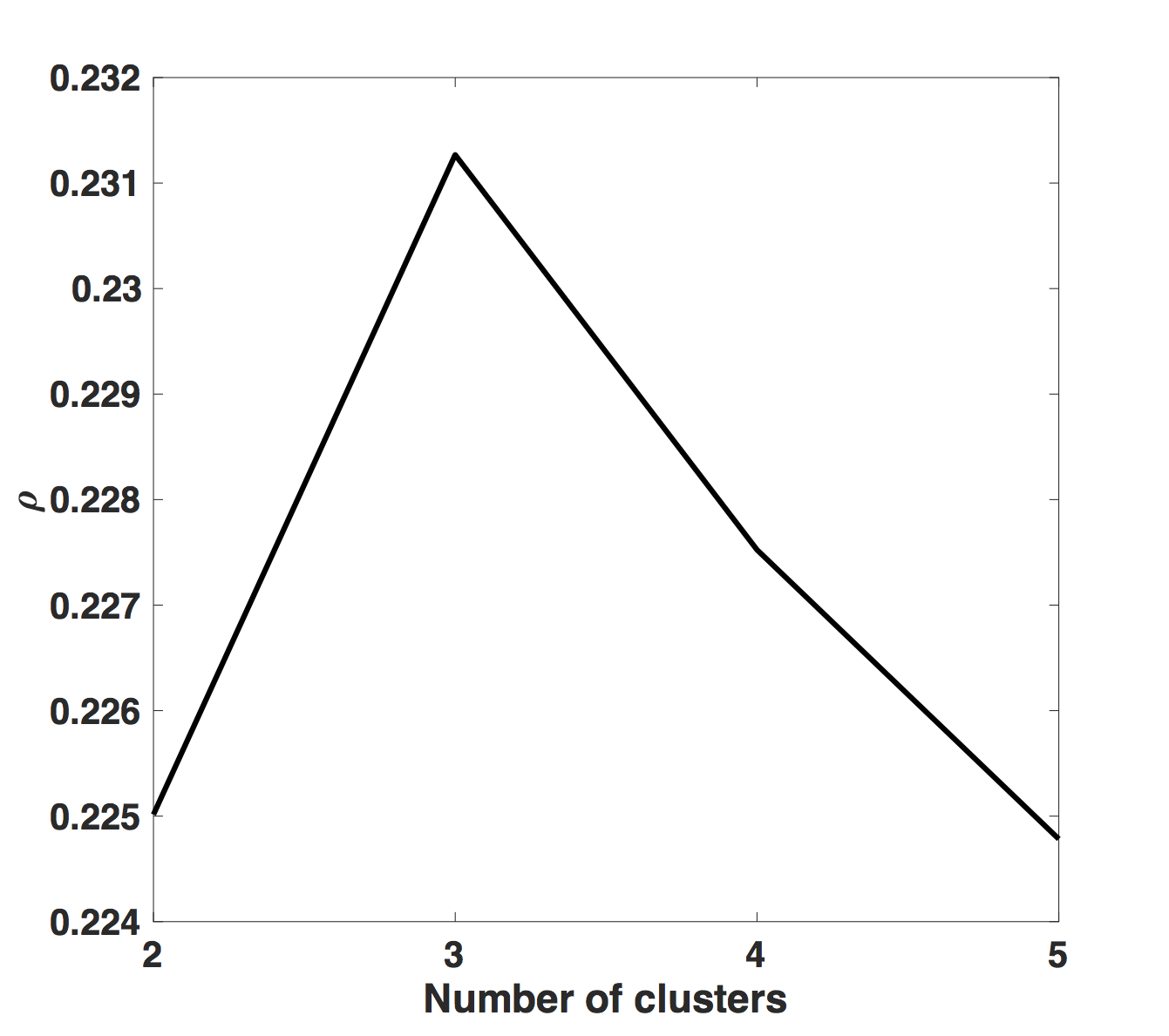}
& \hspace{0pt} 
\end{tabular}}
\caption{Plot of the dispersion coefficient vs. the different number of clusters for the tongue protrusion task. Please note that we analyzed 2-4 clusters in our experiments and the optimal number of clusters was found to be 3 in this dataset.}\label{fig:prot}
\end{figure}
%%%%%%%%%%%%%%%%%

%%%%%%%%%%%%%%%%%%%%%%%%%%%%%%%%%%%%%%%%%%%%%%%%%%%%%%%%
\def\FigureHeight{41mm}
\begin{figure*}[h]
 \center{
 \begin{tabular}{c@{ }c}
   \includegraphics[trim=0mm 0mm 0mm
0mm,clip=true,height=\FigureHeight]{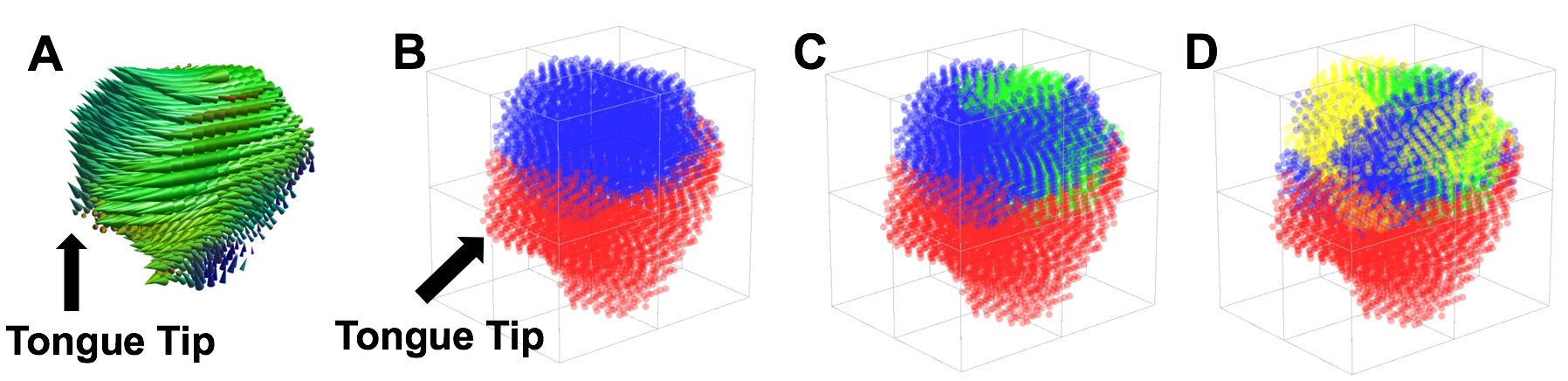}
& \hspace{0pt} 
\end{tabular}}
\caption{Illustration of identified functional units using our method during $/s/$ to $/u/$ from ``a souk," depicting (A) 3D Lagrangian displacement field, (B) detected functional units (2 clusters), (C) detected functional units (3 clusters), and (D) detected functional units (4 clusters). Please note that the displacement fields and clustering results are plotted relative to the first time frame representing the neutral tongue position. The cone size in (A) represents the magnitude of the displacement field, and red, green, and blue represent left--right, front--back, and up--down directions of the displacement field, respectively.}\label{fig:asouk}
\end{figure*}
%%%%%%%%%%%%%%%%%
%%%%%%%%%%%%%%%%%%%%%%%%%%%%%%%%%%%%%%%%%%%%%%%%%%%%%%%%
\def\FigureHeight{75mm}
\begin{figure}[h]
 \center{
 \begin{tabular}{c@{ }c}
   \includegraphics[trim=0mm 0mm 0mm
0mm,clip=true,height=\FigureHeight]{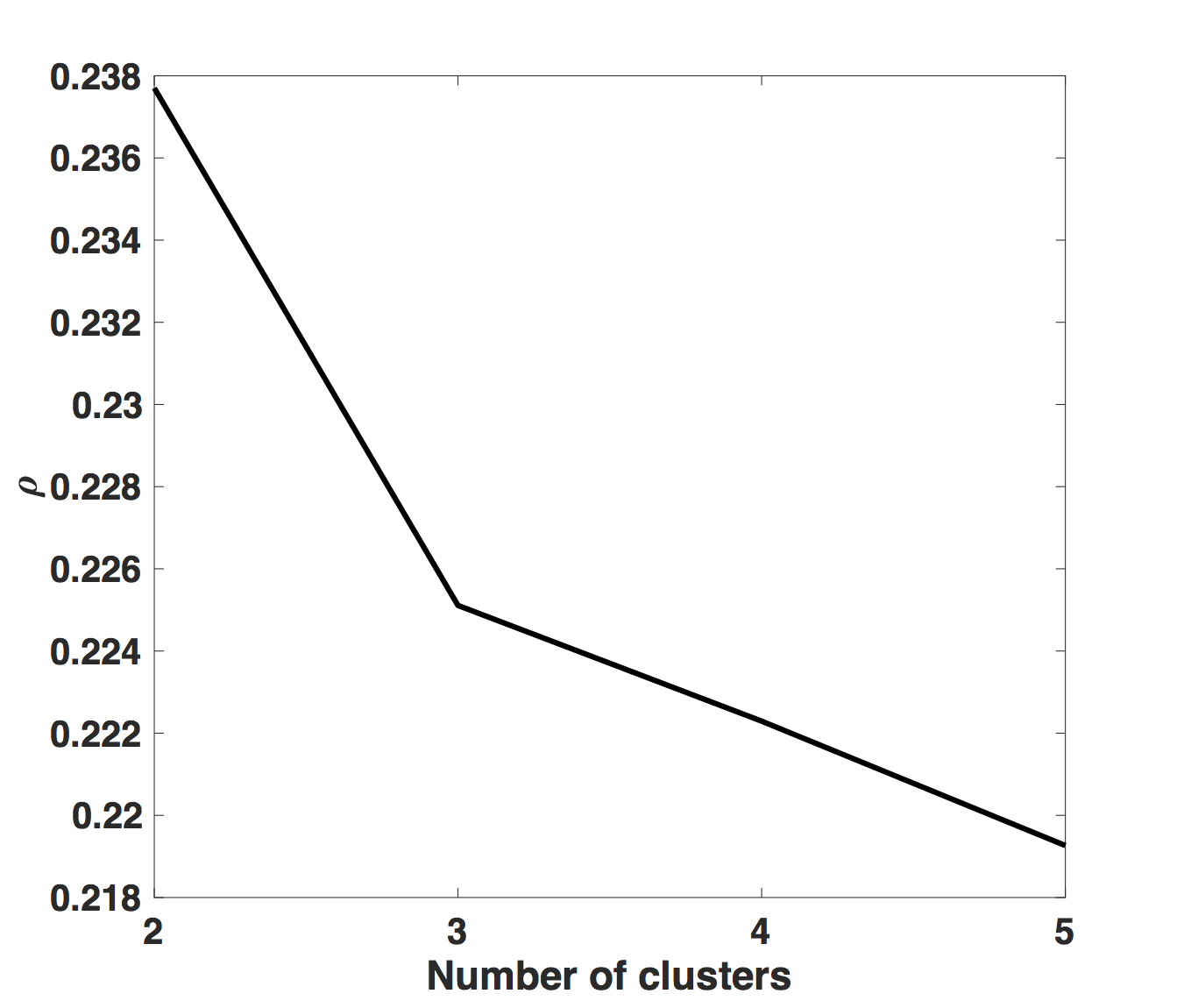}
& \hspace{0pt} 
\end{tabular}}
\caption{Plot of the dispersion coefficient vs. the different number of clusters for the task of $/s/$ to $/u/$ from ``a souk.'' Please note that we analyzed 2-4 clusters in our experiments and the optimal number of clusters was found to be 2 in this dataset.}\label{fig:souk_disp}
\end{figure}
%%%%%%%%%%%%%%%%%

%%%%%%%%%%%%%%%%%%%%%%%%%%%%%%%%%%%%%%%%%%%%%%%%%%%%%%%%
\def\FigureHeight{39mm}
\begin{figure*}[h]
 \center{
 \begin{tabular}{c@{ }c}
   \includegraphics[trim=0mm 0mm 0mm
0mm,clip=true,height=\FigureHeight]{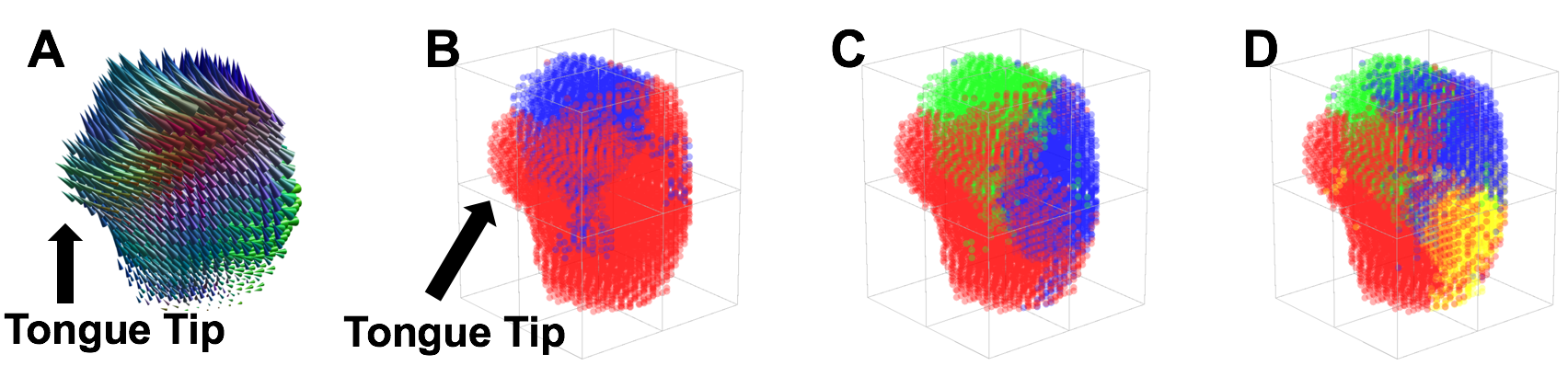}
& \hspace{0pt} 
\end{tabular}}
\caption{Illustration of identified functional units using our method during $/i/$ to $/s/$ from ``a geese," depicting (A) 3D Lagrangian displacement field, (B) detected functional units (2 clusters), (C) detected functional units (3 clusters), and (D) detected functional units (4 clusters). Please note that the displacement fields and clustering results are plotted relative to the first time frame representing the neutral tongue position. The cone size in (A) represents the magnitude of the displacement field, and red, green, and blue represent left--right, front--back, and up--down directions of the displacement field, respectively.}\label{fig:ageese}
\end{figure*}
%%%%%%%%%%%%%%%%%
%%%%%%%%%%%%%%%%%%%%%%%%%%%%%%%%%%%%%%%%%%%%%%%%%%%%%%%%
\def\FigureHeight{75mm}
\begin{figure}[h]
 \center{
 \begin{tabular}{c@{ }c}
   \includegraphics[trim=0mm 0mm 0mm
0mm,clip=true,height=\FigureHeight]{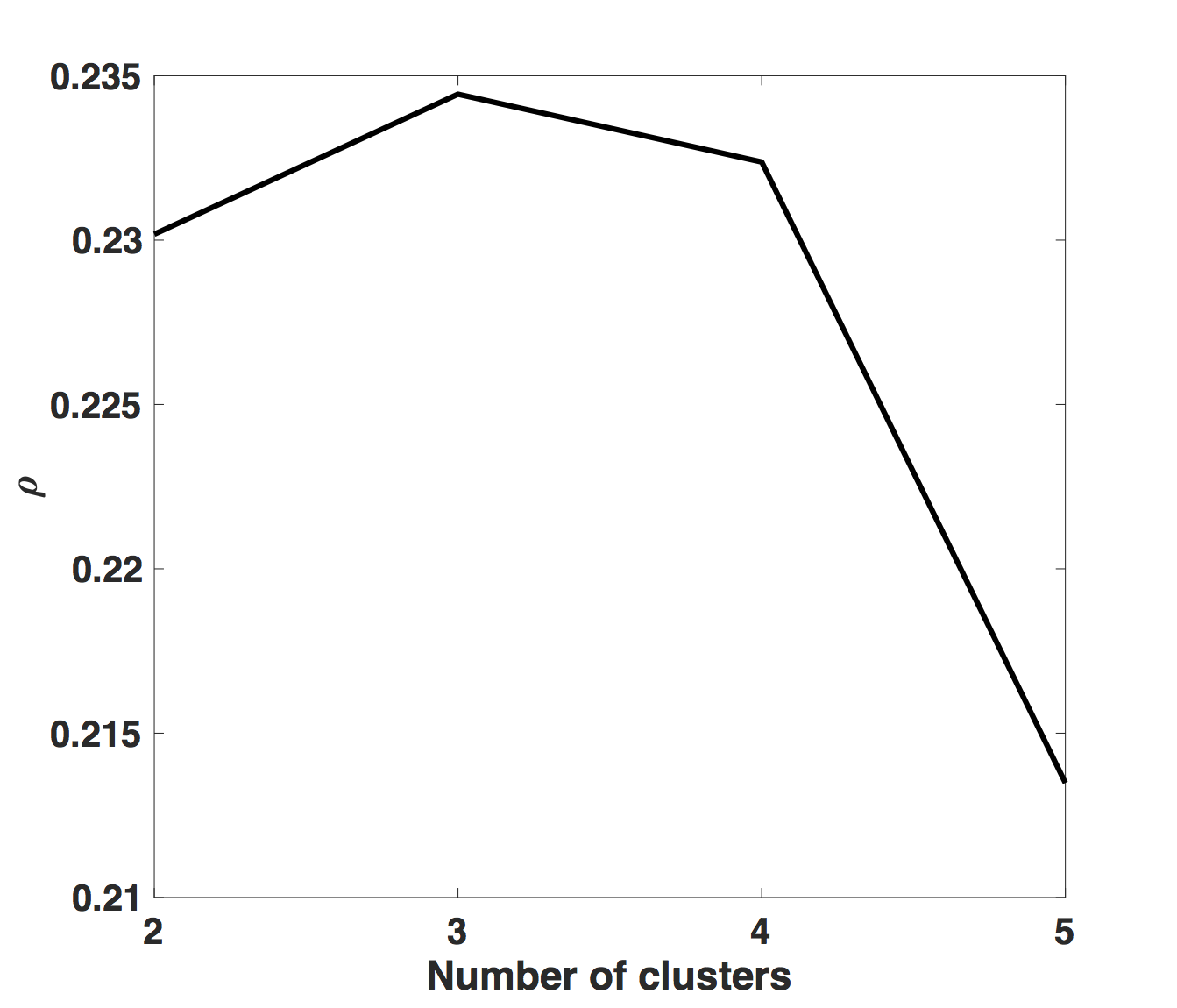}
& \hspace{0pt} 
\end{tabular}}
\caption{Plot of the dispersion coefficient vs. the different number of clusters for the task of $/i/$ to $/s/$ from ``a geese." Please note that we analyzed 2-4 clusters in our experiments and the optimal number of clusters was found to be 3 in this dataset.}\label{fig:geese_disp}
\end{figure}
%%%%%%%%%%%%%%%%%

%% Table 3 %%%%%%%%%%%%%%%%%%%%%%%%%%%%%%%%%%%%%%%%%%%%%%%%%%%%%%%
\begin{table*}[h]
\caption{The size of the detected functional units of /s/-/u/ from ``a souk''}
\centering
%\scriptsize{ 
\begin{tabular}{c||c|c|c|c|c|c|c|c|c} \hline 
 %& 2 units & 3 units & 4 units \\ \hline 
 & \multicolumn{2}{|c|} {2 units} & \multicolumn{3}{|c|} {3 units} & \multicolumn{4}{|c} {4 units}  \\ \hline \hline
 Subject 1 & 47.6$\%$ & 52.4$\%$ & 43.4$\%$ & 19.4$\%$ & 37.2$\%$ & 25.5$\%$ & 33.8$\%$ & 19.1$\%$ & 21.6$\%$ \\ 
 Subject 2 & 46.2$\%$ & 53.8$\%$ & 30.3$\%$ & 18.8$\%$ & 50.9$\%$ & 21.5$\%$ & 36.2$\%$ & 25.3$\%$ & 17.0$\%$ \\
 Subject 3 & 36.3$\%$ & 63.7$\%$ & 28.1$\%$ & 37.5$\%$ & 34.4$\%$ & 25.1$\%$ & 32.1$\%$ & 19.7$\%$ & 23.1$\%$ \\
 Subject 4 & 46.6$\%$ & 53.4$\%$ & 33.3$\%$ & 31.9$\%$ & 34.8$\%$ & 29.0$\%$ & 24.5$\%$ & 22.8$\%$ & 23.7$\%$ \\
 Subject 5 & 49.3$\%$ & 50.7$\%$ & 28.3$\%$ & 41.6$\%$ & 30.1$\%$ & 25.4$\%$ & 27.8$\%$ & 22.0$\%$ & 24.8$\%$ \\
 Subject 6 & 32.1$\%$ & 67.9$\%$ & 37.2$\%$ & 22.6$\%$ & 40.1$\%$ & 22.0$\%$ & 26.5$\%$ & 20.2$\%$ & 31.3$\%$ \\
 Subject 7 & 35.1$\%$ & 64.9$\%$ & 37.6$\%$ & 29.6$\%$ & 32.8$\%$ & 25.2$\%$ & 31.0$\%$ & 20.4$\%$ & 23.4$\%$ \\
 Subject 8 & 36.7$\%$ & 63.3$\%$ & 18.8$\%$ & 46.8$\%$ & 34.3$\%$ & 33.5$\%$ & 21.4$\%$ & 15.3$\%$ & 29.7$\%$ \\ 
 Subject 9 & 42.1$\%$ & 57.9$\%$ & 26.3$\%$ & 35.0$\%$ & 38.8$\%$ & 25.1$\%$ & 24.5$\%$ & 18.3$\%$ & 32.1$\%$ \\
 Subject 10 & 43.4$\%$ & 56.6$\%$ & 47.2$\%$ & 20.4$\%$ & 32.4$\%$ & 28.4$\%$ & 27.6$\%$ & 25.1$\%$ & 18.9$\%$ \\
 \hline \hline
Mean$\pm$SD & 41.5$\pm$6.1$\%$ & 58.5$\pm$6.1$\%$ & 33.0$\pm$8.5$\%$ & 30.4$\pm$9.9$\%$ & 36.6$\pm$5.9$\%$ & 26.1$\pm$3.5$\%$ & 28.5$\pm$4.6$\%$ & 20.8$\pm$3.1$\%$ & 24.6$\pm$5.1$\%$ \\
 \hline 
\end{tabular}\label{table:size_souk}
\end{table*}
%%%%%%%%%%%%%%%%%

%% Table 4 %%%%%%%%%%%%%%%%%%%%%%%%%%%%%%%%%%%%%%%%%%%%%%%%%%%%%%%
\begin{table*}[h]
\caption{The size of the detected functional units of /i/-/s/ from ``a geese''}
\centering
%\scriptsize{ 
\begin{tabular}{c||c|c|c|c|c|c|c|c|c} \hline 
 %& 2 units & 3 units & 4 units \\ \hline 
 & \multicolumn{2}{|c|} {2 units} & \multicolumn{3}{|c|} {3 units} & \multicolumn{4}{|c} {4 units}  \\ \hline \hline
 Subject 1 & 44.2$\%$ & 55.8$\%$ & 38.2$\%$ & 28.2$\%$ & 33.6$\%$ & 34.0$\%$ & 21.8$\%$ & 25.2$\%$ & 19.0$\%$ \\ 
 Subject 2 & 50.1$\%$ & 49.9$\%$ & 33.2$\%$ & 33.0$\%$ & 33.8$\%$ & 31.5$\%$ & 19.9$\%$ & 26.6$\%$ & 22.0$\%$ \\
 Subject 3 & 39.9$\%$ & 60.1$\%$ & 40.3$\%$ & 24.4$\%$ & 24.4$\%$ & 29.3$\%$ & 28.2$\%$ & 25.2$\%$ & 17.3$\%$ \\
 Subject 4 & 52.7$\%$ & 47.3$\%$ & 39.3$\%$ & 30.4$\%$ & 30.2$\%$ & 23.8$\%$ & 23.3$\%$ & 25.5$\%$ & 27.4$\%$ \\
 Subject 5 & 45.7$\%$ & 54.3$\%$ & 38.5$\%$ & 38.5$\%$ & 23.0$\%$ & 33.5$\%$ & 25.7$\%$ & 24.2$\%$ & 16.6$\%$ \\
 Subject 6 & 72.7$\%$ & 27.3$\%$ & 44.0$\%$ & 23.9$\%$ & 32.1$\%$ & 34.4$\%$ & 14.8$\%$ & 27.7$\%$ & 23.1$\%$ \\
 Subject 7 & 50.0$\%$ & 50.0$\%$ & 48.0$\%$ & 20.2$\%$ & 31.8$\%$ & 25.2$\%$ & 32.0$\%$ & 26.1$\%$ & 16.7$\%$ \\
 Subject 8 & 51.1$\%$ & 48.9$\%$ & 37.4$\%$ & 35.3$\%$ & 27.3$\%$ & 18.7$\%$ & 36.2$\%$ & 24.5$\%$ & 20.6$\%$ \\ 
 Subject 9 & 50.8$\%$ & 49.2$\%$ & 43.6$\%$ & 35.0$\%$ & 21.4$\%$ & 31.0$\%$ & 25.6$\%$ & 24.8$\%$ & 18.6$\%$ \\
 Subject 10 & 50.1$\%$ & 49.9$\%$ & 36.7$\%$ & 33.6$\%$ & 29.7$\%$ & 23.1$\%$ & 32.7$\%$ & 23.8$\%$ & 20.4$\%$ \\
 \hline \hline
Mean$\pm$SD & 50.8$\pm$9.2$\%$ & 49.2$\pm$9.2$\%$ & 40.3$\pm$4.4$\%$ & 29.9$\pm$6.1$\%$ & 28.6$\pm$4.7$\%$ & 29.1$\pm$5.4$\%$ & 25.3$\pm$6.4$\%$ & 25.5$\pm$1.1$\%$ & 20.1$\pm$3.6$\%$ \\
 \hline 
\end{tabular}\label{table:size_geese}
\end{table*}
%%%%%%%%%%%%%%%%%

First, we identified the functional units from the protrusion task, where Fig.~\ref{fig:prot_functional_unit} shows (B) two clusters, (C) three clusters, and (D) four clusters, respectively, based on the dispersion coefficient as depicted in Fig.~\ref{fig:prot} and visual assessment. The optimal number of clusters was found to be 3 in this dataset as shown in Fig.~\ref{fig:prot}. The protrusion motion is characterized by forward and upward motion of the tongue as depicted in Fig.~\ref{fig:prot_functional_unit}(A). Fig.~\ref{fig:prot_functional_unit}(B) shows forward protrusion (red) vs. small motion (blue). In addition, as the number of clusters increases as shown in Fig.~\ref{fig:prot_functional_unit}(C) and (D), subdivision of large regions in small motion (blue, Fig.~\ref{fig:prot_functional_unit}(B)) into small functional units was observed.

Second, we identified the functional units during $/s/$ to $/u/$ from ``a souk" of subject 9 as shown in Table~\ref{table:size_souk}. Fig.~\ref{fig:asouk} depicts (B) two clusters, (C) three clusters, and (D) four clusters, respectively in the same manner. The optimal number of clusters was found to be 2 in this dataset as shown in Fig.~\ref{fig:souk_disp}. Table~\ref{table:size_souk} shows the sizes of detected functional units as percentage, where the detected sizes were different from subject to subject and larger variability in the three detected units was observed. During the course of these motions, the tongue tip moves forward to upward/backward, while the tongue body and the posterior tongue move upward and forward, respectively as depicted in Fig.~\ref{fig:asouk}(A). Fig.~\ref{fig:asouk}(B) shows two clusters including the tip plus bottom of the tongue (red) vs. the tongue body. Three clusters in Fig.~\ref{fig:asouk}(C) show a good representation of the tip plus bottom, body and posterior of the tongue and four clusters in Fig.~\ref{fig:asouk}(D) further subdivided the tongue tip and bottom. 

Third, in the same manner, we identified the functional units during $/i/$ to $/s/$ from ``a geese" of subject 8 as shown in Table~\ref{table:size_geese}. Fig.~\ref{fig:ageese} displays (B) two clusters, (C) three clusters, and (D) four clusters, respectively. The optimal number of clusters was found to be 3 in this dataset as shown in Fig.~\ref{fig:geese_disp}. Table~\ref{table:size_geese} shows the sizes of detected functional units as percentage, where the detected sizes were different from subject to subject and larger variability in the two detected units was observed. During the course of these motions, the tongue tip moves upward, while the tongue body moves backward and the posterior tongue moves forward as shown in Fig.~\ref{fig:ageese}(A). Two clusters in Fig.~\ref{fig:ageese}(B) show a division between the tip plus bottom of the tongue (red) vs. the tongue body (blue). Three clusters in Fig.~\ref{fig:ageese}(C) are a good representation of the tip plus body, dorsum, and posterior of the tongue and four clusters as depicted in Fig.~\ref{fig:ageese}(D) subdivided the posterior of the tongue further.

\section{Discussion}

In this work, we presented an approach for characterizing the tongue's multiple functional degrees of freedom, which is critical to understand the tongue's role in speech and other lingual behaviors. This is because the functioning of the tongue in speech or other lingual behaviors entails successful orchestration of the complex system of 3D tongue muscular structures over time. Determining functional units from healthy controls plays an important role in understanding motor control strategy, which in turn could elucidate pathological or adapted motor control strategy when analyzing patient data such as tongue cancer or Amyotrophic Lateral Sclerosis patients. It has been a long-sought problem that many researchers attempted using various techniques. 

Utilizing recent advances in tMRI-based motion tracking and data mining schemes, a new method for identifying functional units from MRI was presented, which opens new windows for a better understanding of the underlying tongue behaviors and for studying speech production. Unsupervised data clustering using sparse NMF is the task of identifying semantically meaningful clusters using a low-dimensional representation from a dataset. Two constraints in addition to the standard NMF were employed to reflect the physiological properties of 4D tongue motion during protrusion and simple speech tasks. Firstly, the sparsity constraint was introduced to capture the simplest and the most optimized weighting map. Sparsity has been one of important properties for phonological theories~\cite{Browman_1995}, and our work attempted to decode this phenomenon within a sparse NMF framework. Secondly, the manifold regularization was added to capture the intrinsic and geometric relationship between motion features. It also allows preserving the geometric structure between motion features, which is particularly important when dealing with tongue motions that lie on low-dimensional non-Euclidean manifolds. 

NMF is favorably considered in this work over other alternative matrix decomposition techniques such as Principal Component Analysis (PCA), Independent Component Analysis (ICA), factor analysis, or sparse coding for identifying functional units of tongue motion. While PCA, for example, is appropriate to analyze kinematic or biomechanical features that exhibit both positive and negative values without the explicit assumption of underlying muscle activity, NMF is well-suited to analyze any signals resulting from muscle activity that are inherently non-negative~\cite{ting_2011}. This, seemingly small, non-negativity constraint of NMF makes a significant difference between PCA and NMF. First, while, in PCA, the building blocks are orthogonal with each other, in NMF, the building blocks are independent, which means no building blocks can be constructed as a linear combination of other building blocks. Second, in NMF, the building blocks are likely to specify the boundaries (or edges) of the features, thereby constructing a convex hull within which all the features can be found~\cite{ting_2011}. The ability of NMF to define the boundaries of the features specifies a subspace in which any possible combinations of functional units lie, thus yielding physiologically interpretable results. Third, NMF learns parts-based representation in the sense that a set of parts, when summed, constitutes the whole features. In contrast, in PCA, the elements of the building blocks and their weighting map can cancel each other out, thus obliterating building blocks' physical meaningfulness.

In the experiments using 2D synthetic data, the purpose of the experiments was to assess the accuracy of the clustering performance given known cluster labels in an unsupervised learning setting as there is no ground truth in the \textit{in vivo} tongue motion data. In the experiments using 3D tongue motion simulation data, the testing motions in Fig.~\ref{fig:synthetic}(A) and (C) included rotations about the styloid process and displacements about the inferosuperior directions, which is consistent with \textit{in vivo} tongue movements. As for the input features, we used the magnitude and angle of point tracks derived from tMRI. More features could be investigated such as those reflecting mechanical properties including principal strains, curvature, minimum-jerk, or two-thirds power law~\cite{feature_1995} or motion descriptors combining those individual features. 

There are a few model selection methods available ranging from heuristic methods to sophisticated probabilistic methods~\cite{Tibshirani_2001,Kulis_2012} for different clustering methods. In this work, we chose the number of clusters based on previous research on lingual coarticulation, which divided the tongue into a small set of regions between two (i.e., tip and body)~\cite{ohman_1967} and five units (i.e., front-to-back segments for the genioglossus, verticalis, and transverse)~\cite{Stone_2004}. Additionally, we built a ``consensus matrix'' from multiple runs for each $k$ and assessed the presence of block structure. As an alternative, one can compare reconstruction errors for different number $k$ or examine the stability (i.e., an agreement between results) from multiple randomly initialized runs for each $k$. Because there is no ground truth in our tongue data, we have used both visual assessment and the model selection approach in which the previous research on coarticulation provided an upper limit of the number of clusters. A thorough analysis of the optimal number of clusters is a subject for future research.

There are a few directions to improve the current work. First, we used a data-driven approach to determine the functional units, which was visually assessed because of the lack of ground truth. This could be improved by further studies using model-based approaches via biomechanical stimulations~\cite{Stavness_2012} or using electromyography~\cite{Pittman_2012} to co-validate our findings. For biomechanical simulations, subject-specific anatomy and the associated weighting map could be input and inverse simulation can then be used to verify the validity of the obtained weighting map. Second, we used the magnitude and orientation of point tracks as our input features. In order to equal the weight of each input feature, we normalized the feature values in the same range. In our future work, we will further study automatic relevance determination methods to model the interactions among these features to yield the best clustering outcome. Finally, the identified functional units as shown in our experimental results may involve multiple regions that correspond to sub-muscle or multiple muscles. Therefore, we will further perform registration~\cite{Woo_reg_2015} of the identified functional units with the muscular anatomy from individual high-resolution MRI, diffusion MRI~\cite{lee_2018}, or a vocal tract atlas~\cite{Woo_2015}, and study the relationship between the functional units and underlying muscular anatomy.

\section{Conclusion}

We have presented a new algorithm to determine local functional units that link muscle activity to surface tongue geometry during protrusion and simple speech tasks. We evaluated the performance of our approach using various simulated and human tongue motion datasets, demonstrating that our method surpassed conventional methods and accurately clustered the human tongue motion. Our results suggest that it is feasible to identify the functional units using a set of motion features, which has great potential in the improvement of diagnosis, treatment, and rehabilitation in patients with speech-related disorders.

% use section* for acknowledgment
\ifCLASSOPTIONcompsoc
  \section*{Acknowledgments}
\else
  % regular IEEE prefers the singular form
  \section*{Acknowledgment}
\fi

This research was in part supported by National Institutes of Health Grants R21DC016047, R00DC012575, R01DC014717, R01CA133015, and P41EB022544. The authors would like to thank Drs. Joseph Perkell and Sidney Fels for their valuable insights and helpful discussions. The authors also thank Sung Ahn for proofreading the text. 

% Can use something like this to put references on a page
% by themselves when using endfloat and the captionsoff option.
\ifCLASSOPTIONcaptionsoff
  \newpage
\fi

% that's all folks
\end{document}